\documentclass[letterpaper, 10 pt, conference]{ieeeconf}
\IEEEoverridecommandlockouts                              
\usepackage{times}
\usepackage{epsfig}
\usepackage{graphicx}
\usepackage{amsmath}
\usepackage{amssymb}
\usepackage{multirow}

\usepackage{graphicx}
\usepackage{epsfig}
\usepackage{epic,eepic}
\usepackage{times}
\usepackage{mathptmx}
\usepackage{amsfonts}
\usepackage{amsmath}
\usepackage{cite}
\usepackage{color}

\usepackage{times}

\definecolor{red}{rgb}{1,0,0}
\definecolor{green}{rgb}{0,1,0}
\definecolor{blue}{rgb}{0,0,1}
\definecolor{violet}{rgb}{1,0,1}
\definecolor{cyan}{cmyk}{1,0,0,0}
\definecolor{magenta}{cmyk}{0,1,0,0}
\definecolor{yellow}{cmyk}{0,0,1,0}

\definecolor{white}{rgb}{1,1,1}

\usepackage{jumoline}

\usepackage{arydshln}

\newcommand{\CO}[1]{}

\newcommand{\CommentOut}[1]{}

 \newcommand{\editage}[1]{}

\newcommand{\FIG}[3]{
\begin{minipage}[b]{#1cm}
\begin{center}
\includegraphics[width=#1cm]{#2}\\
{\scriptsize #3}
\end{center}
\end{minipage}
}

\title{\LARGE
Use of First and Third Person Views for Deep Intersection Classification 
}

\author{Takeda Koji ~~~~ Tanaka Kanji 
\thanks{Our work has been supported in part by JSPS KAKENHI Grant-in-Aid for Scientific Research (C) 26330297, and 17K00361.}
\thanks{K. Tanaka is with Faculty of Engineering, University of Fukui, Japan.
  K. Takeda is with Graduate School of Engineering, University of Fukui, Japan.
{\tt\small tnkknj@u-fukui.ac.jp}}
}

\newcommand{\figA}{
\begin{figure}[t]
\centering
\FIG{6}{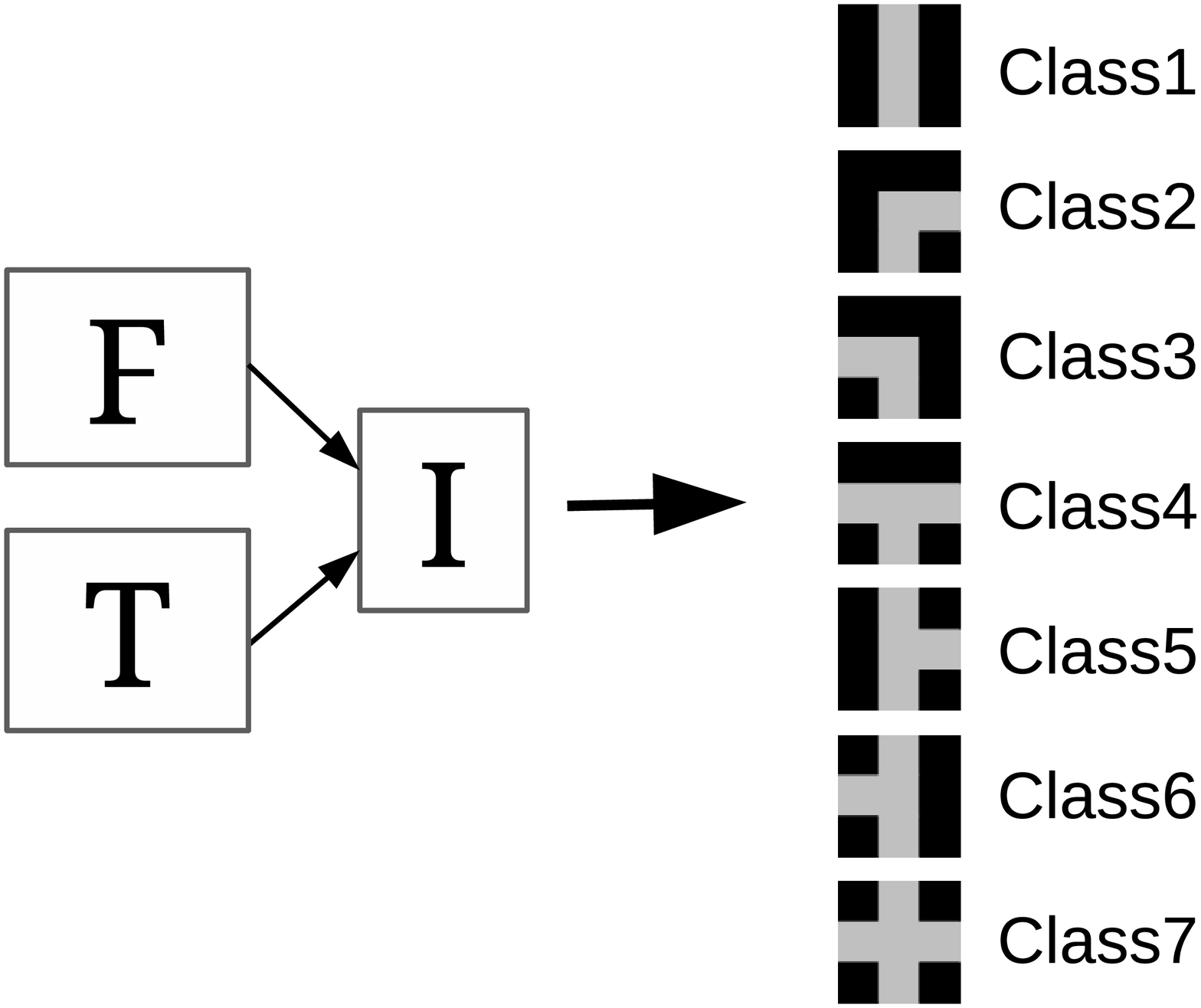}{}
\caption{Overview of the proposed architecture}\label{fig:full_archtecture}
\end{figure}
}

\newcommand{\figB}{
\begin{figure*}[t]
  \centering
  \FIG{4}{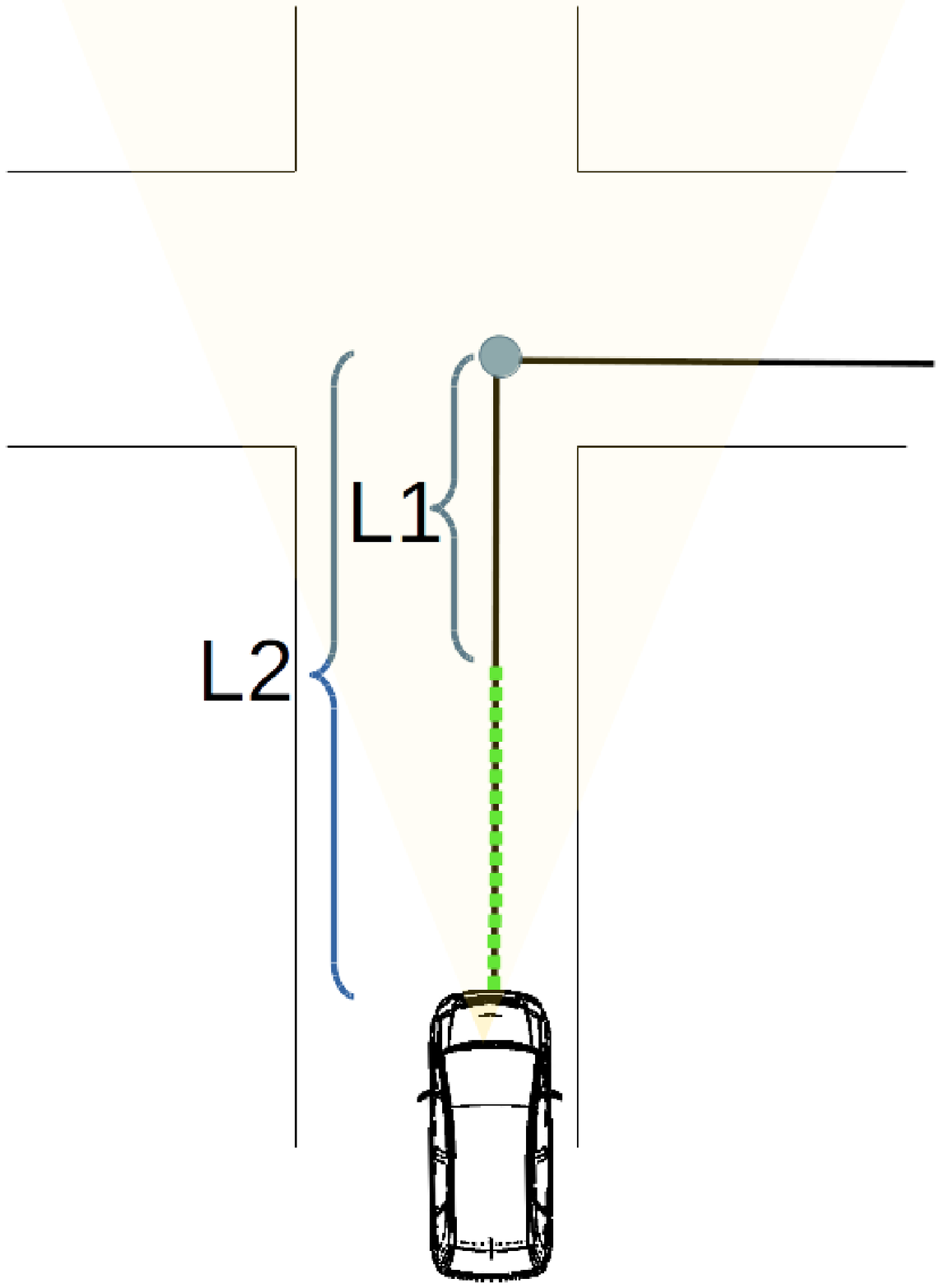}{}\hspace*{-15mm}%
  \FIG{6}{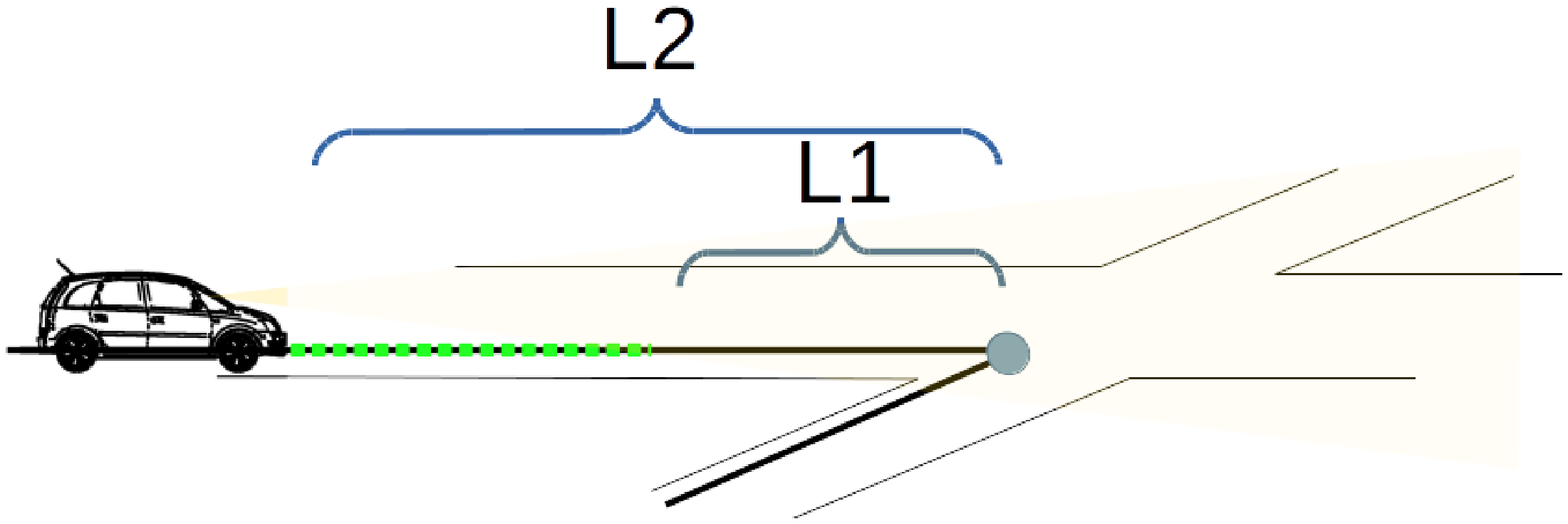}{}\hspace*{-20mm}%
  \FIG{6}{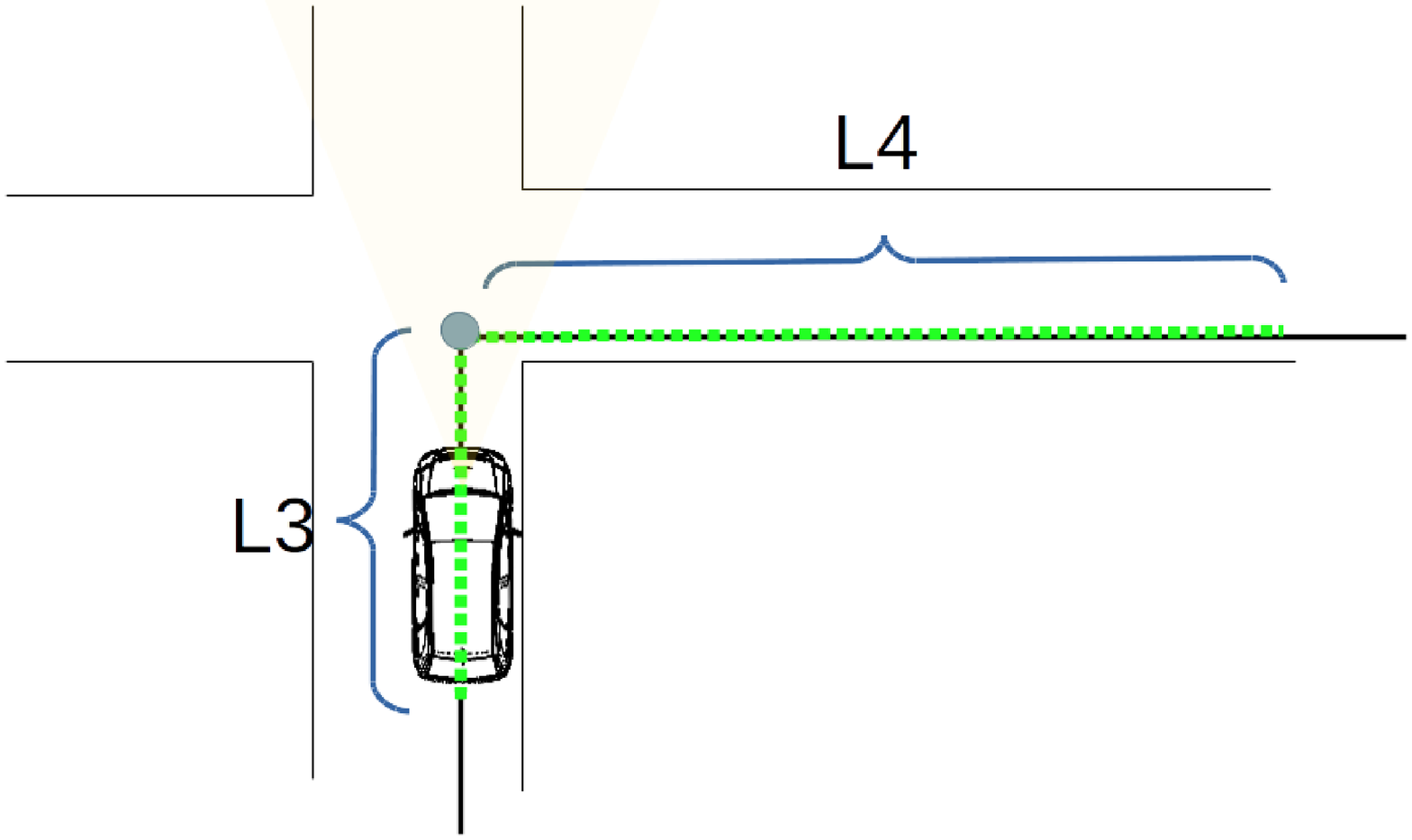}{}\hspace*{-20mm}%
  \FIG{6.5}{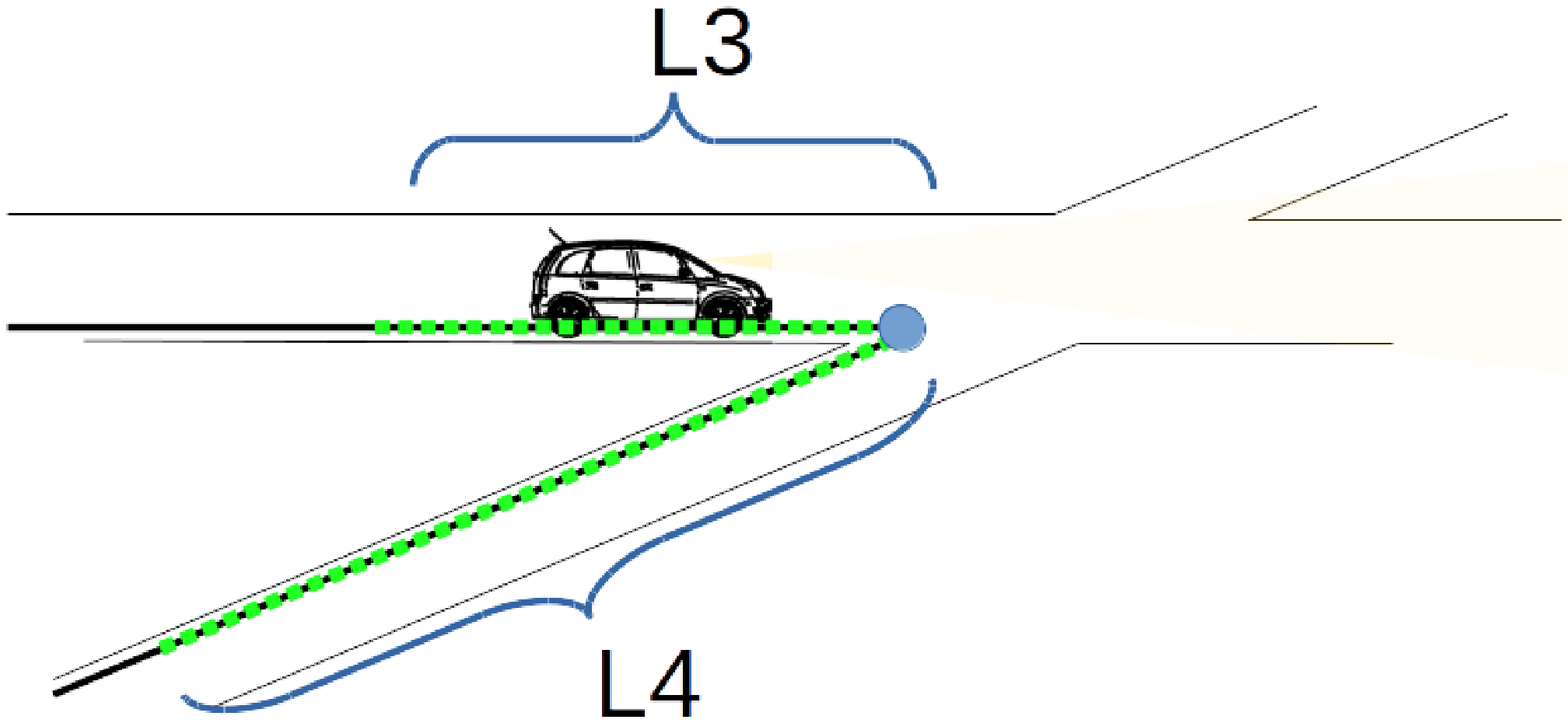}{}
\caption{Distance to intersection (D2I).}
\label{fig:yoko_m}
\end{figure*}
}

\newcommand{\figF}{
\begin{figure}[t]
  \centering
  \vspace*{5cm}
  \FIG{8}{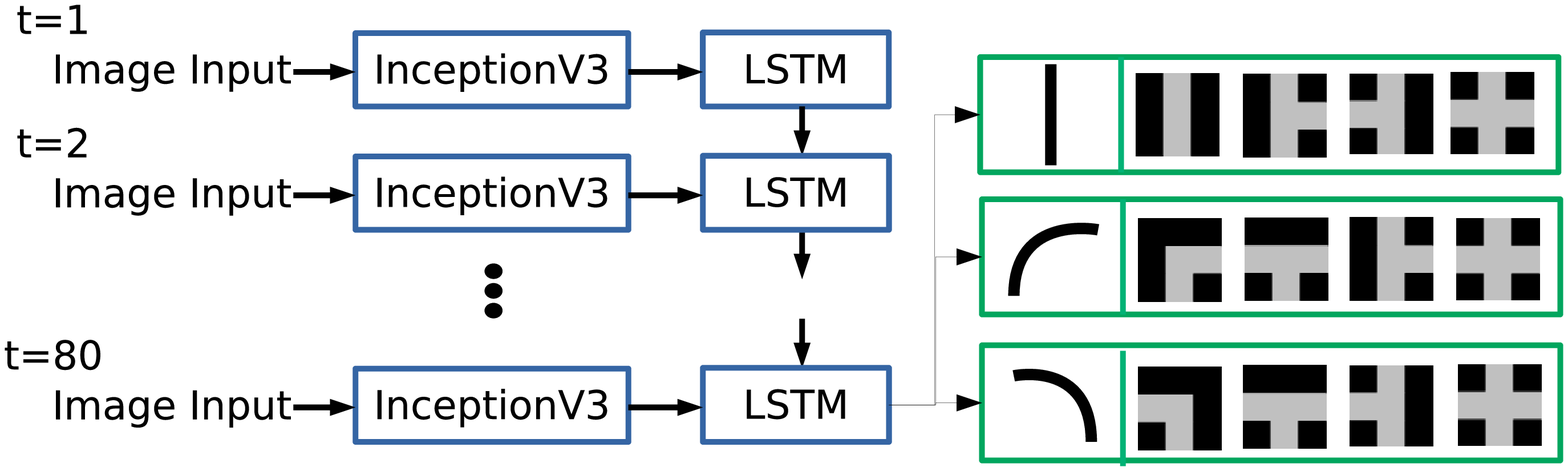}{}
\vspace*{-2cm}
  \caption{F-Net architecture.}\label{fig:snet}
\label{fig:Snetwork}
\end{figure}
}

\newcommand{\figK}{
\begin{figure*}[t]
  \centering
\FIG{8}{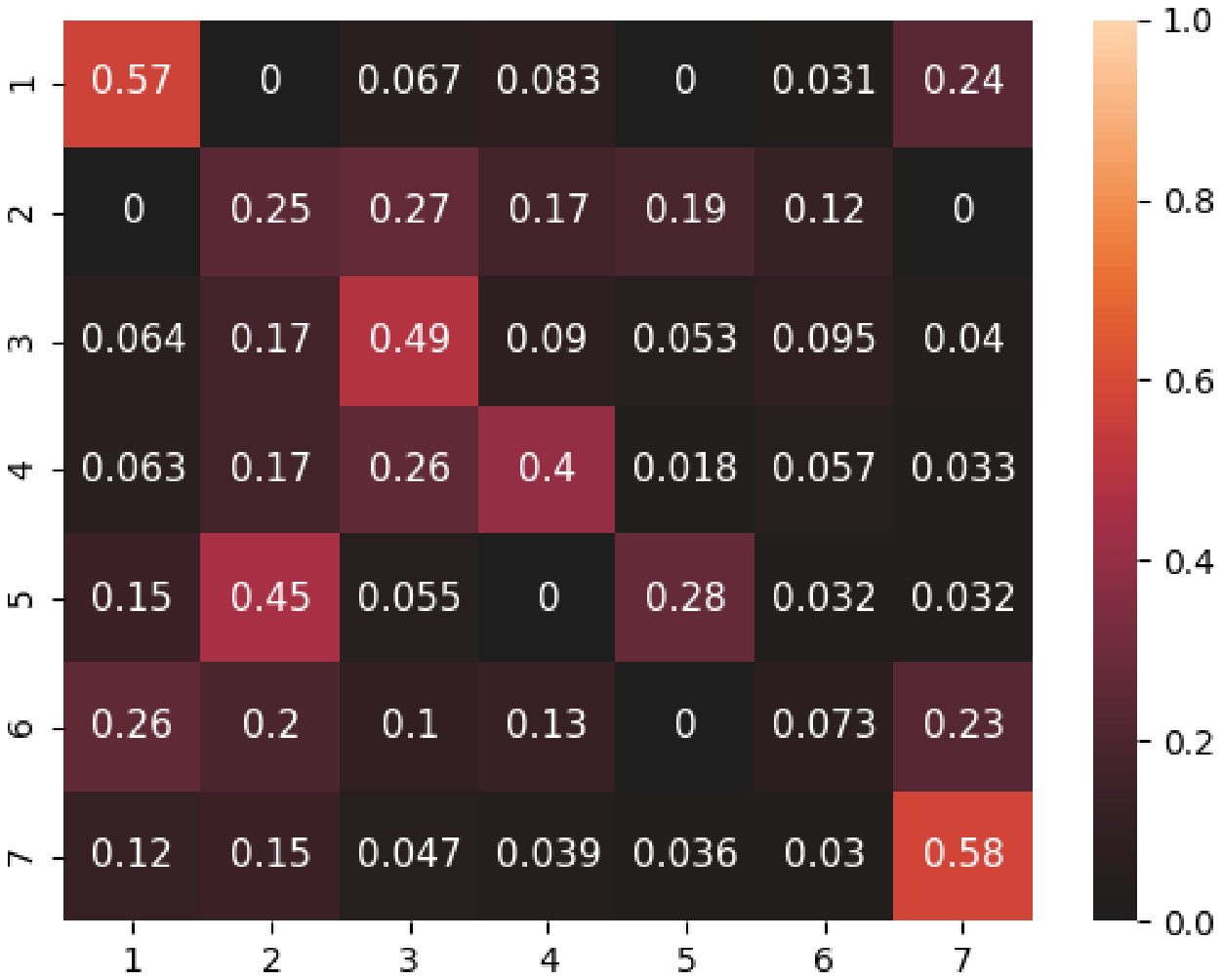}{}
\FIG{8}{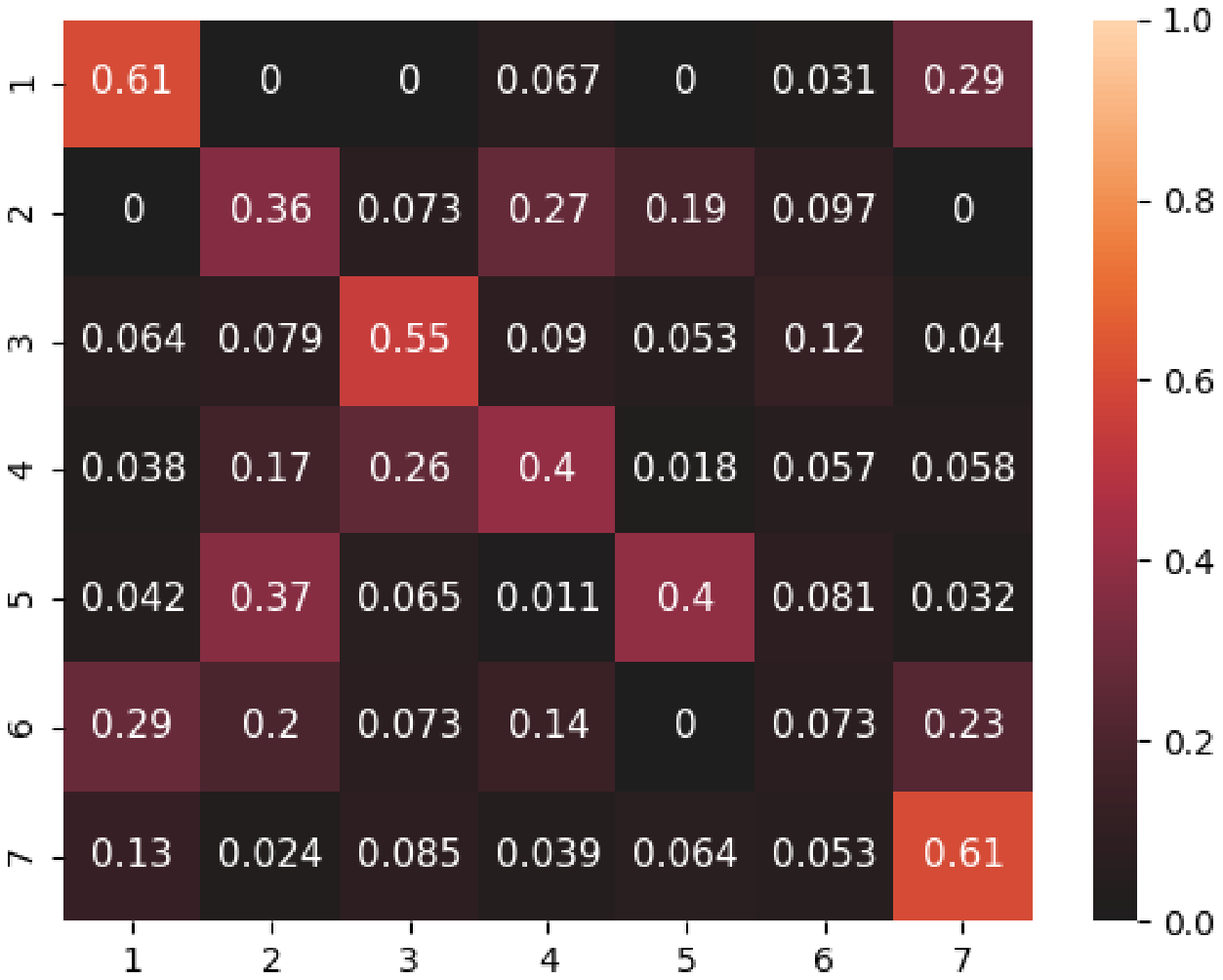}{}
  \caption{Results for the T-Net (left) and the proposed F-T-I-Net (right).}\label{fig:confusion_matrix}
\end{figure*}
}

\newcommand{\figL}{
\begin{table}[t]
\centering
\caption{Results for the performance comparison}\label{table:result}
  \begin{tabular}{|l|r|}
  \hline
Method	& Accuracy \\ \hline
proposed  & 0.42 \\ \hline
T-Net   & 0.37 \\ \hline
  \end{tabular}
\label{table:result}
\end{table}
}

\newcommand{\figN}{
\begin{figure}[t]
\begin{center}
\FIG{12}{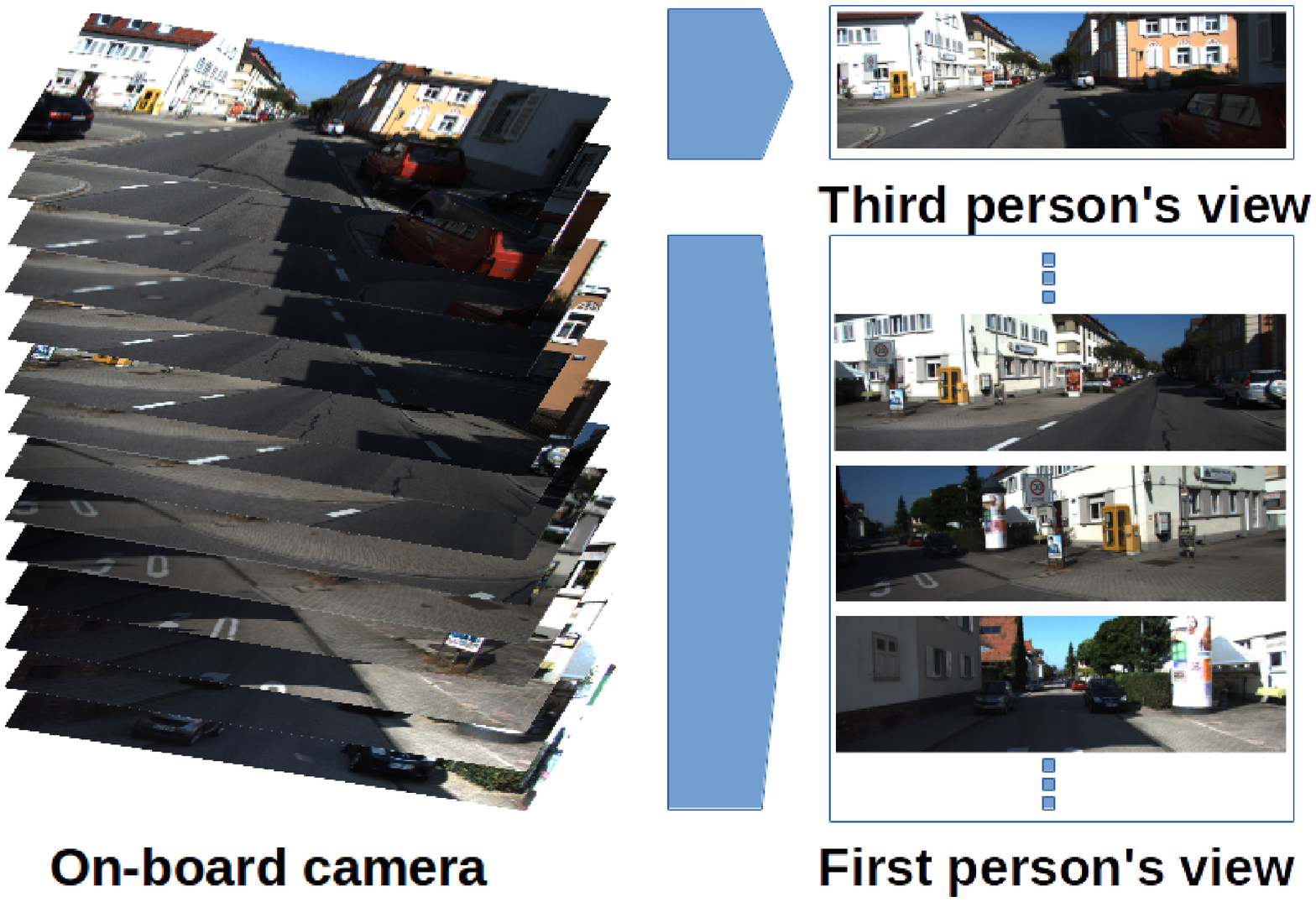}{}
  \caption{%
We combined two independent approaches into a unified framework: the TPV approach, which takes a single view image before entering the intersection (top right), and the FPV approach, which takes an egocentric short-term view sequence while passing the intersection (bottom right).
}\label{fig:N}
\label{fig:compare}
\end{center}
\end{figure}
}

\onecolumn 

\begin{document}

\maketitle

\section*{\centering Abstract}
\textit{
We explore the problem of intersection classification using monocular on-board passive vision, with the goal of classifying traffic scenes with respect to road topology. We divide the existing approaches into two broad categories according to the type of input data: (a) first person vision (FPV) approaches, which use an egocentric view sequence as the intersection is passed; and (b) third person vision (TPV) approaches, which use a single view immediately before entering the intersection. The FPV and TPV approaches each have advantages and disadvantages. Therefore, we aim to combine them into a unified deep learning framework. Experimental results show that the proposed FPV-TPV scheme outperforms previous methods and only requires minimal FPV/TPV measurements.
}

\section{Introduction}

We explored the problem of intersection classification (IC) using monocular on-board passive vision (Fig. \ref{fig:N}),
with the goal of classifying traffic scenes with respect to road topology (e.g., a seven-class problem, as illustrated in
Fig. \ref{fig:full_archtecture}).
This classification could be used to assist drivers directly or to improve autonomous driving (AD). We divided the existing approaches into two broad categories according to the type of input data: (a) first person vision (FPV) approaches, which use a short egocentric view sequence as the intersection is passed; and (b) third person vision (TPV) approaches, which use a single view immediately before entering the intersection.

FPV has been studied
in many different contexts
including 
visual odometry (VO) \cite{DeepVO18}
and lane change detection (LCD) \cite{LaneChangeDetectionNew}.
As shown in Fig. \ref{fig:N},
FPV provides egocentric views characterizing 
the egomotions of 
a vehicle passing an intersection.
These are typically recognized by long-short-term-memory (LSTM) \cite{DeepVO18}.
However,
FPVs are often insufficient to solve the IC problem.
First,
while they are useful for detecting the road 
that the vehicle
{\it did} follow,
they are often useless for detecting the other roads 
that the vehicle {\it did not} follow.
Second,
objects near the intersection may not be invariant visual features 
that can be used for IC.
This may be due to in-motion vibration of the vehicle,
or rapid changes in the relative appearance of the objects.

TPV approaches attempt to solve the IC problem in a different way.
They formulate IC as a scene categorization (SC) problem.
SC is an active research field in AD \cite{SceneCategorizationNew}
and car-robotics \cite{RoboticSceneCategorizationNew}.
As shown in Fig. \ref{fig:N},
TPV provides relatively invariant views 
that can be used to categorize the
relative appearance of the intersection in front of the vehicle \cite{BMVC09IC}.
Furthermore,
its recognition performance can be easily boosted by deep convolutional neural networks (DCN) \cite{ITSC17IC}.
However, 
SC is essentially an ill-posed problem,
and even deep-learning solutions are far from perfect.
Moreover,
in principle 
it is difficult for a passive vision
to recognize
the traversability of roads
at a distance \cite{Traversibility}.

\figN

In this study,
we investigated 
how FPV and TPV 
could be deployed
within a conceptually simple framework for IC.
FPV and TPV approaches 
both have their own advantages and disadvantages.
Hence,
we aimed to combine them into a unified IC framework.
Specifically,
we aimed to
develop a unified deep-learning (DL) framework for IC that
could be used to train two different deep neural networks (DNNs),
LSTM and DCN,
for two different subtasks: FPV and TPV,
which could them be integrated into an intersection classifier.
The distance from the intersection is greater in 
TPV than in FPV,
and the distance to the intersection (D2I) can be measured by 
VO \cite{libviso} or egomotion regression \cite{DeepVO18}.
One unique characteristic of the proposed architecture
is that
it uses two different modal inputs
taken from different viewpoints
at different times.
The experimental results demonstrated that
the proposed FPV-TPV framework outperforms
the previous method
despite 
the fact that it requires 
minimal FPV/TPV measurements.

The proposed FPV and TPV approaches
can be also viewed as
two of the most primitive components 
in visual simultaneous localization and mapping (SLAM),
in which the goal is to estimate the state of the vehicle and its surroundings
from
egocentric motion measurements (i.e., FPV)
and
landmark perception measurements (i.e., TPV).

Our approach infers the intersection class directly
from the FPV/TPV measurements.
Thus,
it is orthogonal to
indirect approaches
that rely on 
intermediate representations,
such as
road segmentation \cite{DBLP:conf/itsc/SuleymanovA018},
road/obstacle reconstruction \cite{DBLP:conf/itsc/NolteKM18},
tracking pedestrians \cite{DBLP:conf/itsc/RidelRLSW18},
traffic signs \cite{DBLP:conf/itsc/GuptaC18}, TLs \cite{DBLP:conf/itsc/FernandezGSS18},
GPS \cite{gps2016itsc}, and map \cite{DBLP:conf/itsc/FladeNVE18}.

All of these previous
approaches
have their own limitations,
such as limited field of views,
texture misrecognition,
unavailability of global information,
complex motions of dynamic objects,
and loss of tracking due to occlusions.
Therefore,
the proposed FPV-TPV approach could be combined with them to improve its classification performance further.

\section{Approach}\label{sec:approach}

We formulated IC as a multi-modal classification problem.
For simplicity, we considered seven typical intersection classes as shown in Fig. \ref{fig:full_archtecture}.
Our IC system takes
two types of input,
input-F and input-T (Fig. \ref{fig:yoko_m}),
and aims to
classify the traffic scene into one of the seven classes.
Input-F is a short view sequence that is taken as the intersection is passed (i.e., the D2I is short).
Input-T is a single view image 
that is taken just before entering the intersection (i.e., the D2I is long).

\figA

The proposed architecture consists of two kinds of base networks: F-Net and T-Net
which take input-F and input-T, respectively.
This is followed by an additional integrator network: I-Net,
which integrates the outputs from F-Net and T-Net
into a result
in the form of a 7-dim probability distribution function (PDF).

F-Net addresses an alternative three-class IC problem,
differentiating between the three classes defined in Fig. \ref{fig:snet}.
The three-class classifier takes input-F
and outputs a 3-dim PDF for the three-class problem.

It is worth nothing
that
implementing F-Net as a seven-class classifier
is a natural choice,
but it turned out
to provide poor performance,
as we will show in Section \ref{sec:snet}.

T-Net addresses the seven-class IC problem,
as does the 
the proposed F-T-I-Net.
Therefore, we use T-Net as one of baseline methods
(Section \ref{sec:exp}).
It takes input-T and outputs the 7-dim PDF for the seven-class problem.

To find the D2I, we used the practical assumption that D2I cannot be precisely measured (by either VO or SLAM), and instead the D2I must be in a pre-defined range. Specifically, the input-T viewpoint was assumed to be in the range $[-L_2, -L_1]$, while the beginning and the end of the input-F view sequence were assumed to be in the ranges $[-L_3, 0]$ and $[0, L_4]$ respectively, as shown in Fig. \ref{fig:yoko_m}. There were two motivations for this, it is useful for (1) deploying in real-world applications as well as (2) for investigating the influence of the D2I on the recognition performance of T-Net and F-Net.

I-Net was designed as a seven-class classifier. It takes the outputs from T-Net and F-Net, and returns a 7-dim PDF for the seven-class problem. One challenge in the design was how to deal with the difference between the D2I in T-Net and F-Net. As mentioned previously, D2I was significantly larger in T-Net (e.g., 10 m) than in F-Net (e.g., 0 m).

The following subsections provide details of the methods used to implement these networks.

\figB
\figF

\subsection{F-Net}\label{sec:snet}

We built the F-Net on LSTM architecture \cite{LstmOrg}. Fig. \ref{fig:Snetwork} illustrates the F-Net architecture. We converted all of the raw training/testing view-sequences into equal-length sequences 80 s long (corresponding to an average travelling distance of 22 m), by using zero-padding (w/ pre-padding).

Initially, we planned simply to use the raw view sequences as an input for the F-Net. We performed preliminary experiments exploring this idea on an independent dataset. Disappointingly, this did not provide good recognition results.

Next, we introduced two additional image channels for optical flow (OF) and modified the F-Net to take the OF-channels instead of raw red green blue (RGB)-channels. To convert a raw RGB image to an OF-image, we applied the OF extraction algorithm in \cite{TEX_URL1}. Then, we computed the OF vector $(u,v)$  for each point in the $u$-$v$ image coordinate, before color-coding the OF vectors. Next, an OF-image was converted to a 2048-dim feature vector, which was then used as the input training/testing data for the F-Net. For this conversion, we employed a pre-trained InceptionV3 net, and used its last intermediate layer (i.e., output of the pooling layer) 2048-dim signal as the feature vector. For training, we used 35 samples of feature vector sequences and set the LSTM and FCL layer of the F-Net as trainable. 
As a result, the performance can be improved by using the OF-based F-Net. Therefore, we used the OF-based F-Net as the default method in our experimental system.

\subsection{T-Net}

We built the T-Net on VGG16-Net architecture \cite{Vgg16Org}. The original VGG-Net aims to train a 1000-class classifier using the ImageNet training set. We modified the last fully connected layer (FCL) of the VGG16-Net from 1000 to 7 in order to deal with the 7-class problem. We then replaced the FCL at the classification layer with two layers consisting of a 512-dim FCL and a 128-dim FCL. At the fine-tuning stage, we froze the layers between the first and fourth max-pooling layers in the VGG16-Net and set the other layers as trainable.

\subsection{I-Net}

Our initial idea was to build the I-Net on a fully connected network (FCN) that would take outputs from the F-Net and T-Net. Based on this, we implemented an I-Net that consisted of three FCLs with dimension 512-128-7. However, the preliminary experiments produced disappointing results. This may be because the I-Net training requires a sufficiently large number of training samples. Since the input format of I-Net (i.e., the outputs from T-Net and F-Net) are unique, it is difficult to apply a standard data augmentation technique to increase the sample set size. As a result, the I-Net may overfit to the training data.

Next, we developed a modified version of I-Net. Instead of using the data-demanding FCN, we derived a multimodal information fusion \cite{EarlyLateFusionNew}, and augmented the T-Net's  PDF of the T-Net using the PDF of the F-Net as a prior in a Bayesian fashion. Suppose we are given a pair of PDFs, the 7-dim PDF $T_{out}$ and the 3-dim PDF in $F_{out}$, from T-Net and  F-Net, respectively. The goal is to integrate them into a more reliable 7-dim PDF vector $I_{out}$. The 3-dim PDF vector $F_{out}$ is interpreted as a prior 7-dim PDF $W[c]$, and the final output of I-Net is computed by 
$I_{out}[c]=T_{out}[c] W[c]$,  where $c$ has the class ID 
($c\in [1,7]$).
In this study, $W$ was modeled as a binary mask vector.
Each $c$-th element $W[c]$ is set 1 
if the class $c$ is not inconsistent
with the worst-1 class $c^*$ of $F_{out}$, 
otherwise it is 0.
In addition,
when
$F_{out}[c^*]$ for the top-1 class $c^*$
is greater than 0.999,
each $c$-th element $W[c]$ is set 1 if the class $c$ is consistent
with the top-1 class $c^*$ of $F_{out}$, otherwise it is 0.
Although
we used a binary formulation for
the Bayesian multimodal information fusion,
it could be extended
to a real-value
Bayesian formulation
by learning the prior PDF
in a data-driven manner.

\section{Evaluation Experiments}\label{sec:exp}

This section describes our experiments.
We augmented the KITTI dataset \cite{KITTI} with ground-truth annotation. 
The KITTI dataset is an outdoor dataset acquired entirely
in the City of Karlsruhe, Germany. 
The original image size was 1,241$\times$376. 
We used the left hand images from the on-board stereo camera as an input for our IC system. 
We resized images to 224$\times$224 for the T-Net input, 
and 299$\times$299 for the F-Net input.

We made comparisons between the proposed method and 
the fully-deep T-Net (presented in Section \ref{sec:approach}).

The training procedures were as follows.
A GPU (GeForce GTX 1080 Ti) was used for training the T-Net and F-Net. For programming,
Keras with the tensorflow backend was used. 
For both networks, the learning rate was initialized at $1.0\times 10^{-5}$, and the weight decay was set to $1.0\times 10^{-6}$. 
We used an Adam optimizer \cite{AdapOptimizerOrg}.
To suppress overfitting, we introduce dropout layers and early stopping.
We analyzed GPS data, which was provided as a part of the KITTI dataset, and manually collected
410
input-T images for the seven different intersection classes.
The number of images in each class was 42, 46, 46, 79, 43, 72, and 82 for class \#1, \#2, \#3, \#4, \#5, \#6, and \#7, respectively.
We divided the entire data set into training, validation, and testing sets, with a ratio of 5:2:3
for both input-T and input-F.
The number of input-F sequences was 10 for each class and thus 70 in total. For performance evaluation, 5-fold cross validations were used. For the cross validation, we ensured that the test and validation sets consisted of novel intersections that were not seen in the training stage. Since every training/testing sample for the I-Net consisted of an input-T and input-F pair, and the number of available input-F samples was smaller than the input-T sample, input-F samples could appear multiple times in the sample set.

We use the performance metric of top-1 accuracy, which is defined as the ratio of successfully classified test samples that are top-ranked by a classifier.

Table \ref{table:result} summarizes the results of the proposed method and the comparison methods.
It is clear that the proposed method outperformed 
the comparison T-Net method. 
Fig. \ref{fig:confusion_matrix} shows confusion matrices for the T-Net and the proposed F-T-I-Net. The proposed approach outperformed the T-Net for almost all of the seven classes.
Specifically,
the proposed method performed almost the same as the T-Net for class-4 and class-6,
and is significantly better for the other classes.

\section{Conclusions}

We have presented a method for IC that combines first and third person views taken from different viewpoints at different points in time. Our method produces a model that is more robust than previous IC methods. We employed a single-view scene categorization formulation for TPV based on DCN, and a view sequence classification formulation for FPV based on LSTM. Experimental results showed that the proposed FPV-TPV framework outperformed the previous approach despite the fact that it required minimal FPV/TPV measurements.

\figK
\figL

\bibliographystyle{IEEEtran}
\bibliography{cite}

\end{document}